%% file: CGnet.tex
\documentclass[11pt]{amsart}
\pdfoutput=1
\usepackage{iclr2018_conference,times}
\usepackage{url,amsthm,amsmath,amssymb,amsfonts}
\usepackage{xcolor}
\usepackage{algorithm,algpseudocode}
\usepackage{tikz}
\usepackage{bbm}
\usepackage{rbase} 
\usepackage{paralist}
\definecolor{hcitecolor}{RGB}{40,40,160}
\usepackage[colorlinks=true,citecolor=hcitecolor, linkcolor=black]{hyperref}
\usepackage{caption}
\usepackage{subfigure}
\usepackage{graphicx}
\usepackage{pgfplots}
\usepackage{xypic}
\usepackage[all,2cell,ps,cmtip]{xy}
\usepackage{tikz}
\usepackage{caption}
\usepackage{mdframed}

\algdef{SE}[DOWHILE]{Do}{doWhile}{\algorithmicdo}[1]{\algorithmicwhile\ #1}%
\algnewcommand{\varalg}{\textit}

\title{\m{N}--body networks: a covariant hierarchical neural network architecture for 
learning atomic potentials\m{{}^{1}}}

\usepackage{tikz}
\usetikzlibrary{shadows}
\usetikzlibrary{matrix, calc, positioning, arrows,shapes,backgrounds, arrows.meta, quotes}
\usepackage{gensymb,stmaryrd,mathtools,textcomp,xspace,grffile,rbase}

\newtheorem{theorem}{Theorem}

\newtheorem{proposition}[theorem]{Proposition}

\newtheorem{definition}{Definition}

\ignore{
\makeatletter
\def\thm@space@setup{%
  \thm@preskip=6pt plus 0pt minus 0pt
  \thm@postskip=0pt plus 0pt minus 0pt 
}
\makeatother
}

\newcommand{\includepic}[2]{\includegraphics[width=#1\textwidth]{#2}}

\newcommand{\neuron}{\mathfrak{n}}

\newcommand{\prt}{\Pcal}
\newcommand{\ch}{\textrm{ch}}

\newcommand{\sm}[1]{\m{\smash{#1}}}

\newcommand{\risi}[1]{{\color{blue}[Risi: #1]}}

\iclrfinalcopy 

\begin{document}

\maketitle\footnote{This note describes the neural network architecture first presented by the author at 
the ``Machine Learning for Molecules and Materials'' workshop at the Neural Information Processing Systems 
Conference (Long Beach, CA) on December 8, 2017.}

\vspace{-40pt}
\textbf{Risi Kondor}\\ 
Departments of Computer Science \& Statistics\\
The University of Chicago\\
\texttt{risi@cs.uchicago.edu}
\\ \\ 

\begin{abstract}
\input{abstract}

\end{abstract}

\input{intro}
\input{nn}
\input{phys}
\input{activations}
\input{conclusions} 
\input{acknowledgements}

\bibliography{CGnet}
\bibliographystyle{iclr2018_conference}

\end{document}

%% file: abstract.tex
We describe \m{N}--body networks, a neural network architecture for learning the behavior and properties 
of complex many body physical systems. 
Our specific application is to learn atomic potential energy surfaces for use in molecular dynamics simulations. 
Our architecture is novel in that (a) it is based on a hierarchical decomposition of the many body system 
into subsytems (b) the activations of the network correspond to the internal state of each subsystem 
(c) the ``neurons'' in the network are constructed explicitly so as to guarantee that each of the 
activations is covariant to rotations (d) the neurons operate entirely in Fourier space, and the 
nonlinearities are realized by tensor products followed by Clebsch--Gordan decompositions. 
As part of the description of our network, we give a characterization of what way the weights of the 
network may interact with the activations so as to ensure that the covariance property is maintained.

%% file: intro.tex
\section{Introduction}

In principle, quantum mechanics provides a perfect description of the forces governing the 
behavior of atomic systems such as crystals and biological molecules. 
However, for systems larger than a few dozen atoms, solving  
the Schr\"odinger equation explicitly, on present day computers, is not a feasible proposition. 
Even Density Functional Theory (DFT) \citep{HohenbergKohn}, 
a widely used approximation in quantum 
chemistry, has trouble scaling to more than about a hundred atoms. 

Consequently, the majority of practical work in molecular dynamics 
foregoes modeling electrons explicitly, and 
falls back on the fundamentally classical (i.e., non-quantum) 
Born--Oppenheimer approximation, which 
treats atoms as solid balls that exert forces on nearby balls prescribed by so-called (effective)  
atomic potentials. 
Assume that the potential attached to atom \m{i} is \m{\phi_i(\sseq{\h {\V r}}{k})},  
with \m{\h{\V r_j}=\V r_{p_j}\!\<-\V r_i}, where \m{\V r_i} is the position vector of atom \m{i} and 
\m{\V r_{p_j}} is the position vector of its \m{j}'th neighbor.  
The total force experienced by atom \m{i} is then simply the negative gradient  
\m{F_i=-\nabla_{\!\small \V r_i} \phi_i(\sseq{\h {\V r}}{k})}. 
Classically, in molecular dynamics \m{\phi_i} is usually given in terms of a closed form formula with a few tunable 
parameters. Popular examples of such so-called empirical potentials (empirical force fields) 
include the CHARMM models \citep{CHARMM1983,CHARMM2009} and others. 
 
Empirical potentials are fast to evaluate but are crude 
models of the quantum interactions between atoms, 
limiting the accuracy of molecular simulation. 
A little over ten years ago, machine learning entered this field, promising 
to bridge the gap between the quantum and classical worlds by \emph{learning} 
the aggregate force on each atom as a function of the positons of its neighbors from a relatively 
small number of DFT calculations \citep{Behler2007fe}. 
In the last few years there has been a veritable explosion in the amount of activity in 
machine learned atomic potentials (MLAP), 
and molecular dynamics simulations based on this approach 
are starting to yield results that outperform other methods  
\citep{Bartok2010wd,Behler2015gv,Shapeev2015,Chmiela2016a,Zhang2017,Schutt2017}.  

Much of the arsenal of present day machine learning algorithms has been applied to the MLAP problem,  
from genetic algorithms, through kernel methods, to neural networks. 
However, rather than the statistical 
details of the specific learning algorithm, often what is critically important for 
problems of this type 
is the representation of the atomic environment, i.e., the choice of learning features  
that the algorithm is based on. 
This situation is by no means unique in the world of applied machine learning:  
in computer vision and speech recognition, in particular, there is a rich literature 
of such representational issues. 
What makes the situation in Physics applications somewhat special is the presence of 
constraints and invariances that the representation must satisfy not just in an approximate, 
but in the \emph{exact} sense.  
As an example, one might consider rotation invariance. 
If rotation invariance is not fully respected by an image recognition system, some objects 
might be less likely to be accurately detected in certain orientations than in others. 
In a molecular dynamics setting, however, using a potential that is not fully rotationally invariant 
would not just degrade accuracy, but would likely lead to entirely unphysical molecular trajectories.  

\subsection{Fixed vs.\:learned representations.} 
Similarly to other branches of machine learning, in recent years the MLAP community has been shifting from 
fixed input features towards representations learned from the data itself, 
in particular, using ``deep'' neural networks to represent atomic enviroments. 
Several authors have found that certain concepts from the mainstream neural networks literature,  
such as convolution and equivariance, can be successfuly repurposed to this domain. 
In fact, the analogy with computer vision is more than just skin deep. 
In both domains two competing objectives are critical to success: 
\begin{compactenum}[~~1.]
\item 
The ability to capture structure in the input data at multiple different length scales, 
, i.e., to construct a \emph{multiscale} representation of the input image or the atomic environment. 
\item 
The above mentioned invariance property with respect to spatial transformations, including  
translations, rotations, and possibly scaling. 
\end{compactenum}
There is a rich body of work on addressing these objectives in the neural networks 
literature. One particularly attractive approach is 
the \emph{scattering networks} framework of Mallat and coworkers, 
which, at least in the limit of an infinite number of neural network layers, 
provides a representation of functions that is both 
globally invariant with respect to symmetries and Lipschitz with respect to warpings 
\citep{Mallat2012,Hirn2017}. 

Inspired by recent work on neural networks for representing graphs and other structured objects 
by covariant compositional neural architectures \citep{CompNetsArxiv18}, 
in this paper we take the idea of learnable multiscale representations one step further,  
and propose \m{N}--body networks, 
a neural network architecture  where \emph{the individual ``neurons'' 
correspond to physical subsystems endowed with their own internal state}.
The structure and behavior of the resulting model 
follows the tradition of coarse graining and representation theoretic ideas in Physics, and 
provides a learnable and multiscale representation of the atomic environment that is fully 
covariant to the action of the appropriate symmetries. 
However, the scope of the underlying ideas is significantly broader, and we believe that \m{N}--body networks 
will also find application in modeling other types of many-body Physical systems, as well. 

An even more general contribution of the present work is that it shows how the machinery of group 
representation theory, specifically the concept of Clebsch--Gordan decompositions, can be used to design 
neural networks that are covariant to the action of a compact group yet are computationally efficient. 
This aspect is related to the recent explosion of interest in generalizing the notion of convolutions to graphs 
\citep{Niepert2016,Defferrard2016,Duvenaud2015,Li2016,Riley2017,CompNetsArxiv18}, 
manifolds \citep{Monti2016, Masci2015}, 
and other domains \citep{Bruna2013,SphericalCNN2018},  
as well as the question of generalizing the concept of equivariance (covariance) in general   
\citep{Cohen2016,Cohen2017,EquivarianceArxiv18}. 
Several of the above works employed generalized Fourier representations of one type or another, 
but to ensure equivariance the nonlinearity was always applied in the ``time domain''. Projecting 
back and forth between the time domain and the frequency domain is a major bottleneck, 
which we can eliminate because the Clebsch--Gordan 
transform allows us to compute one type of nonlinearity, tensor products, entirely in the Fourier domain.



%% file: nn.tex
\section{Representing structured objects with neural nets}\label{sec: nn}

To put our work in perspective, we begin with reviewing classical feed-forward neural networks, 
and then describe a relatively new, general purpose neural architecture for representing structured 
objects called \emph{compositional networks}.  

A prototypical feed-forward neural network consists of some number of 
neurons \m{\cbrN{\neuron^\ell_i}} arranged in \m{L\<+1} distinct layers. 
Layer \m{\ell\<=0} is the \emph{input layer}, where training and testing data enter the network, 
while the inputs of the neurons 
in layers \m{\ell=1,2,\ldots,L} are the outputs 
\m{\cbrN{f^{\ell-1}_j}} of the neurons in the previous layer. Each neuron 
computes its output (also called its \emph{activation}) using a simple rule such as 
\begin{equation}\label{eq: classical aggregation}
f^\ell_i=\sigma\brBig{\sum_j w^\ell_j\, f^{\ell-1}_j+b_\ell}, 
\end{equation}
where the \m{\cbrN{w^\ell_j}} weights and \m{\cbrN{b_\ell}} biases are learnable parameters, 
while \m{\sigma} is a fixed nonlinearity, such a sigmoid function or a ReLU operator.  
The output of the network appears in layer \m{L}, is compared with the desired output by means 
of a loss function, and the gradient of the loss is back-propagated through the network 
to update the parameters, usually by some variant of stochastic gradient descent.  

One of the reasons commonly cited for the spectacular success of feed-forward 
neural networks (especially ``deep'', i.e., many layer ones) 
is their ability to implicitly decompose complex objects into their constituent parts. 
This is especially true of \emph{convolutional} neural networks (CNNs), 
commonly used in computer vision \citep{LeCun1998}. 
In CNNs, the weights in each layer are tied together, which tends to   
force the neurons to learn increasingly complex visual features, from simple edge detectors 
all the way to complex shapes such as human eyes, mouths, faces, and so on. 

\subsection{Compositional networks} 
There has been a lot of interest in extending neural networks to learning 
from structured objects, such as graphs.
A range of architectures have been proposed for this purpose, many of them based on 
various generalizations of the notion of convolution to these domains 
\citep{Duvenaud2015,Kearns2016,Niepert2016,Riley2017}.

One particular architecture, which makes the part-based aspect of neural modeling very explicit, 
is that of \emph{compositional networks} (comp-nets), introduced in \citep{CompNetsArxiv18}. 
To represent a structured object \m{\Xcal}, comp-nets start with decomposing \m{\Xcal} into 
a hierarchy of parts, subparts, sub-subparts, and so on, 
down to some number of elementary parts \m{\cbr{e_i}}, forming a so-called 
\emph{composition scheme}\footnote{ 
Note that in \citep{CompNetsArxiv18} the elementary parts are called \emph{atoms}, but we will avoid this 
terminology to avoid possible confusion with the physical meaning of the word.}. 
Since each part \m{\prt_i} can be a sub-part of more than one higher level part, 
the composition scheme is not necessarrily a tree, 
but is rather a DAG (directed acyclic graph), as in Figure \ref{fig: parts}. 
\input{fig-parts}
The exact definition is as follows.  

\begin{definition}
Let \m{\Xcal} be a compound object with \m{n} elementary parts \m{\Ecal=\cbrN{\sseq{e}{n}}}. 
A \df{composition scheme} \m{\Dcal} for \m{\Xcal} is a directed acyclic graph (DAG) 
in which each node \m{\neuron_i} is associated with some subset \m{\prt_i} of \m{\Ecal}  
(these subsets are called the \df{parts} of \m{\Xcal}) in such a way that 
\begin{compactenum}[~1.]
\item If \m{\neuron_i} is a leaf node, then \m{\prt_i} contains a single elementary part \m{e_{\xi(i)}}. 
\item \m{\Dcal} has a unique root node \m{\neuron_r}, which corresponds to the entire set \m{\cbrN{\sseq{e}{n}}}. 
\item For any two nodes \m{\neuron_i} and \m{\neuron_j}, 
if \m{\neuron_i} is a descendant of \m{\neuron_j}, then \m{\prt_i\subset \prt_j}. 
\end{compactenum}
\end{definition}

\noindent 
A comp-net is essentially just a  composition scheme reinterpreted as a feed-forward neural network.  
In particular, in a comp-net each ``neuron'' \m{\neuron_i} also has an activation \m{f_i}. 
For leaf nodes, \m{f_i} is some simple pre-defined vector representation of the corresponding 
elementary part \m{e_{\xi(i)}}.  
For internal nodes, \m{f_i} is computed from the activations 
\m{f_{\ch_1},\ldots,f_{\ch_k}} of the children of \m{\neuron_i} by the use of some aggregation function 
\m{\Phi(f_{\ch_1},\ldots,f_{\ch_k})} similar to \rf{eq: classical aggregation}.  
Finally, the output of the comp-net is the output of the root node \m{\neuron_r}. 

\citet{CompNetsArxiv18} discuss in detail the behavior of comp-nets under transformations of 
\m{\Xcal}, in particular, how to ensure that the output of the network is invariant with respect 
to spurious permutations of the elementary parts, 
whilst retaining as much information about the combinatorial structure of \m{\Xcal} as possible. 
This is especially important in graph learning, 
the original problem that motivated the introduction of comp-nets,  
where \m{\Xcal} is a graph, \m{\sseq{e}{n}} are its vertices, and \m{\cbrN{\prt_i}} are subgraphs 
of different radii. 
The proposed solution, \emph{covariant compositional networks (CCNs)}, involves turning the 
\m{\cbrN{f_i}} activations into \emph{tensors} that transform in prescribed ways with respect to 
permutations of the elementary parts making up each \m{\prt_i}. 

\ignore{
\subsection{Neural models for physical systems}

In Physics there is a long history of decomposing systems into a hierarchy of interacting 
subsytems at different scales, from coarse graining approaches to renormalization group theory \risi{cite}. 
Such a multiscale or multiresoltion approach is especially 
well suited to our problem of representing atomic environments. 
For example, to calculate the aggregate force on a central atom, in a first approximation one would 
just sum up independent contributions from each of its neighbors. In a second approximation, however, 
one would consider the modifying effect of the local neighborhoods of the neighbors, and so 
on.

The framework of covariant compositional networks is thus a natural starting point for 
learning atomic force fields, or aggregate properties of physical systems, in general. 
However, physical problems have a number of distinguishing features that 
set them apart from, e.g., learning properties of graphs:
\begin{compactenum}[~~CM1.]
\item 
To learn the properties of many-particle systems, such as the force exerted on 
a central atom by a collection of atoms in its neighborhood, it is natural to take the elementary 
parts of the comp-net to be the constituent atoms, in the physical sense of the word.  
At higher levels in the network, however, identifying subsystems explicitly with  subsets of atoms might 
be too restrictive. After all, chemistry is all about atoms interacting via their shared electron clouds, 
which quantum mechanics describes by wave functions that involve \emph{all} participating 
particles. 
\item 
Physical processes always play out in some space \m{\Omega}, in the simplest case, \m{\Omega\<=\RR^3}.  
In learning force fields, each atom \m{p_i} is naturally associated with a point 
\m{\x_i\tin \Omega}, which is the location of its nucleus. 
To describe interactions between subsystems, however, it will be advantageous to also associate a 
point \m{\x_j} with higher level nodes of the network. 
Our general framework allows us to consider spaces more general than just \m{\RR^3}, but 
we will always assume that \m{\Omega} has vector space strucutre. 
\item
A further departure from idealized combinatorial objects is that physical systems have internal states. 
Whatever is computed by a given node of our generalized compositional model is a 
representation of the state of the corresponding subsystem. 
Therefore, to make this change in perspective explicit and emphasize that each node in our forthcoming 
models has physical reality, instead of talking about \m{f_i} 
activations, we endow each node with a \emph{state space} \m{V_i}, and regard the quantity 
computed at that node as a \emph{state vector} \m{\phi_i\tin V_i}. 
\item 
Finally, and critically, we need to ensure the appropriate behavior of the model with respect to the action 
of symmetry groups. In the simplest case, the group \m{G} is a transformation group of \m{\Omega} itself, 
such as the group \m{\SO(3)} of rotations of \m{\RR^3}. 
However, generally, \m{G} will act on not just the \m{\x_i} 
position vectors, but also on the internal states. 
Similary to quantum mechanics, we assume that this action is linear, which implies that 
each \m{V_i} is a \emph{representation space} of \m{G} 
in the algebraic sense of the word, and we denote the corresponding (reducible or irreducible) 
representation by \m{\rho_i}. For background in representation theory, the reader is referred 
to the Appendix. 
\end{compactenum} 
In summary, we have the following generalization of covariant compositional networks. 

\begin{definition}\label{def: cnm}
Let \m{\mathcal S} be a physical system that lives in a space \m{\Omega} acted on by a 
symmetry group \m{G}. 
A \df{covariant neural model} \m{\Mcal} for \m{\Scal} consists of 
a directed acyclic graph \m{\Dcal} in which 
\begin{compactenum}[~~1.]
\item Each node \m{\neuron_j} is associated with a point \m{\x_j\tin\Omega}, a representation space 
\m{V_j} of \m{G}, and a state vector \m{\phi_j\tin V_j}. 
\item If \m{\neuron_j} is a non-leaf node and its children are \m{\neuron_{\ch_1},\ldots,\neuron_{\ch_k}}, 
then \m{\phi_j} is expressible as a function 
\[\Phi(\V r_{\ch_1},\ldots,\V r_{\ch_k},\phi_{\ch_1}\ldots,\phi_{\ch_k}), \]
where \m{\V r_{\ch_{i}}=\x_{\ch_i}-\x_j}. 
\item Under the action of G on \m{\Scal}, each \m{\phi_j} state vector transforms according to the 
correponding representation \m{\rho_j}.  
\end{compactenum}
If \m{\Dcal} has a unique root \m{\neuron_r}, then the learned state of the entire system \m{\Scal} 
is \m{\phi_r}. 
\end{definition}
\bigskip

\noindent 
When \m{\Scal} is made up of \m{n} distinct particles the above specializes as follows. 

\begin{definition}\label{def: mbnn}
If \m{\Scal} is a multi-particle system consisting of \m{n} particles \m{\sseq{p}{n}}, 
then each leaf node \m{\neuron_i} of the associated neural model \m{\Mcal} corresponds to a single 
particle \m{p_{\ell(i)}}, in particular, \m{\x_{\ell(i)}} is the 
position vector of  \m{p_{\ell(i)}} and \m{\phi_{\ell_i}} is the vector describing \m{p_{\ell(i)}}'s internal 
state. The resulting model we call a \df{multi-body neural network} (MBNN).    
\end{definition}
\bigskip
}





%% file: fig-parts.tex
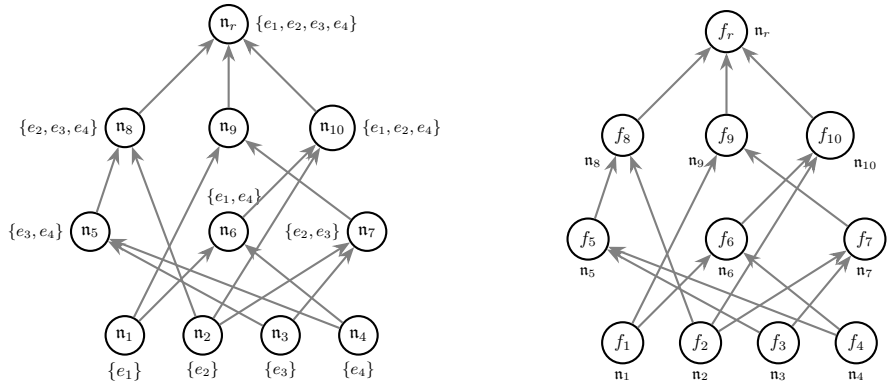
\begin{figure}[t!]
\begin{center}
	
\tikzset{cnodesmall/.style={circle,draw,thick,minimum size=.25em}}

\begin{tikzpicture}[scale = 0.23, every node/.style={scale=0.7},level/.style={},decoration={brace,mirror,amplitude=7}, >={Stealth[gray]}]

 
\node (0dot1) [cnodesmall,draw=black,thick,label={below: $\{e_1\}$}] at (-6,-4) {$\textcolor{black}{\mathfrak{n}_{1}}$};
\node (0dot2) [cnodesmall,draw=black,thick,label={below: \small $\{e_2\}$}] at (-1.5,-4) {$\textcolor{black}{\mathfrak{n}_{2}}$};
\node (0dot3) [cnodesmall,draw=black,thick,label={below: \small $\{e_3\}$}] at (3,-4) {$\textcolor{black}{\mathfrak{n}_{3}}$};
\node (0dot4) [cnodesmall,draw=black,thick,label={below: \small $\{e_4\}$}] at (7.5,-4) {$\textcolor{black}{\mathfrak{n}_{4}}$};



\node (prdot1) [cnodesmall,draw=black,thick,label={left: \small $\{e_3, e_4\}$}] at (-8,2) {$\textcolor{black}{\mathfrak{n}_{5}}$}; 
\node (prdot2) [cnodesmall,draw=black,thick,label={above: \small $\phantom{L}\{e_1, e_4\}$}] at (0,2) {$\textcolor{black}{\mathfrak{n}_{6}}$};
\node (prdot3) [cnodesmall,draw=black,thick,label={left: \small $\{e_2, e_3\}$}] at (8,2) {$\textcolor{black}{\mathfrak{n}_{7}}$};

\node (trdot1) [cnodesmall,draw=black,thick,label={left: \small $\{e_2, e_3, e_4\}$}] at (-6,8) {$\textcolor{black}{\mathfrak{n}_{8}}$}; 
\node (trdot2) [cnodesmall,draw=black,thick,label={}] at (0,8) {$\textcolor{black}{\mathfrak{n}_{9}}$}; 
\node (trdot3) [cnodesmall,draw=black,thick,label={right: \small $\{e_1, e_2, e_4\}$}] at (6,8) {$\textcolor{black}{\mathfrak{n}_{10}}$}; 

\node (qrdot1) [cnodesmall,draw=black,thick,label={right: \small $\{e_1, e_2, e_3, e_4\}$}] at (0,14) {$\textcolor{black}{ \mathfrak{n}_r}$};

\draw[thick,->, gray,fill=gray] (0dot3) -- (prdot1); \draw[thick,->, gray] (0dot4) -- (prdot1); 
\draw[thick,->, gray] (0dot1) -- (prdot2); \draw[thick,->, gray] (0dot4) -- (prdot2);
\draw[thick,->, gray] (0dot2) -- (prdot3); \draw[thick,->, gray] (0dot3) -- (prdot3); 

\draw[thick,->, gray, fill=gray] (0dot2) -- (trdot1); \draw[thick,->, gray] (prdot1) -- (trdot1);
\draw[thick,->, gray] (0dot2) -- (trdot3); \draw[thick,->, gray] (prdot2) -- (trdot3); 
\draw[thick,->, gray] (0dot1) -- (trdot2); \draw[thick,->, gray] (prdot3) -- (trdot2);

\draw[thick,->, gray] (trdot1) -- (qrdot1); \draw[thick,->, gray] (trdot2) -- (qrdot1);
\draw[thick,->, gray] (trdot3) -- (qrdot1);

\end{tikzpicture}	
\hspace{40pt}
\begin{tikzpicture} [scale=0.23,every node/.style={scale=0.7}, level/.style={},decoration={brace,mirror,amplitude=7}, >={Stealth[gray]}]


\node (0dot1) [cnodesmall,draw=black,thick,label={below: \small $\textcolor{black}{\mathfrak{n}_{1}}$}] at (-6,-4) {$f_1$};
\node (0dot2) [cnodesmall,draw=black,thick,label={below: \small $\textcolor{black}{\mathfrak{n}_{2}}$}] at (-1.5,-4) {$f_2$};
\node (0dot3) [cnodesmall,draw=black,thick,label={below: \small $\textcolor{black}{\mathfrak{n}_{3}}$}] at (3,-4) {$f_3$};
\node (0dot4) [cnodesmall,draw=black,thick,label={below: \small $\textcolor{black}{\mathfrak{n}_{4}}$}] at (7.5,-4) {$f_4$};



\node (prdot1) [cnodesmall,draw=black,thick,label={below: \small $\textcolor{black}{\mathfrak{n}_{5}}$}] at (-8,2) {$f_{5}$}; 
\node (prdot2) [cnodesmall,draw=black,thick,label={below: \small $\textcolor{black}{\mathfrak{n}_{6}}$}] at (0,2) {$f_{6}$};
\node (prdot3) [cnodesmall,draw=black,thick,label={below: \small $\textcolor{black}{\mathfrak{n}_{7}}$}] at (8,2) {$f_{7}$};

\node (trdot1) [cnodesmall,draw=black,thick,label={below left: \small $\textcolor{black}{\mathfrak{n}_{8}}$}] at (-6,8) {$f_8$}; 
\node (trdot2) [cnodesmall,draw=black,thick,label={below left: \small $\textcolor{black}{\mathfrak{n}_{9}}$}] at (0,8) {$f_9$}; 
\node (trdot3) [cnodesmall,draw=black,thick,label={below right: \small $\textcolor{black}{\mathfrak{n}_{10}}$}] at (6,8) {$f_{10}$}; 

\node (qrdot1) [cnodesmall,draw=black,thick,label={right: \small $\textcolor{black}{\mathfrak{n}_{r}}$}] at (0,14) {$\textcolor{black}{f_r}$};

\draw[thick,->, gray] (0dot3) -- (prdot1); \draw[thick,->, gray] (0dot4) -- (prdot1); 
\draw[thick,->, gray] (0dot1) -- (prdot2); \draw[thick,->, gray] (0dot4) -- (prdot2);
\draw[thick,->, gray] (0dot2) -- (prdot3); \draw[thick,->, gray] (0dot3) -- (prdot3); 

\draw[thick,->, gray] (0dot2) -- (trdot1); \draw[thick,->, gray] (prdot1) -- (trdot1);
\draw[thick,->, gray] (0dot2) -- (trdot3); \draw[thick,->, gray] (prdot2) -- (trdot3); 
\draw[thick,->, gray] (0dot1) -- (trdot2); \draw[thick,->, gray] (prdot3) -- (trdot2);

\draw[thick,->, gray] (trdot1) -- (qrdot1); \draw[thick,->, gray] (trdot2) -- (qrdot1);
\draw[thick,->, gray] (trdot3) -- (qrdot1); 
   
\end{tikzpicture}
\end{center}
\caption{\label{fig: parts}
(a) A \emph{composition scheme} for an object \m{\Xcal} is a DAG in which the leaves correspond to 
the elementary parts of \m{\Xcal}, 
the internal nodes correspond to sets of elementary parts, and the root corresponds to the entire object.  
(b) A \emph{compositional network} is a composition scheme in which each node \m{\neuron_i} 
also carries a feature vector (activation) \m{f_i}, which is computed 
from the feature vectors of the children of \m{\neuron_i}. 
}
\end{figure}

%% file: phys.tex
\section{Compositional models for atomic environments}

\noindent 
Decomposing complex systems into a hierarchy of interacting subsytems at different scales is a recurring 
theme in physics, from coarse graining approaches to renormalization group theory.  
The same approach applied to the atomic neighborhood lends itself naturally to learning force fields. 
For example, to calculate the aggregate force on the central atom, in a first approximation one 
might just sum up independent contributions from each of its neighbors. In a second approximation, 
one would also consider the modifying effect of the local neighborhoods of the neighbors. 
A third order approximation would involve considering the neighborhoods of the atoms in these 
neighborhoods, and so on. 

The compositional networks formalism is thus a natural framework for force field learning. 
In particular, we consider comp-nets in which the elementary parts correspond to actual physical 
atoms, the internal nodes correspond to subsystems \m{\prt_i} made up of multiple atoms, and 
the corresponding activation, which we now denote \m{\psi_i}, and call the \emph{state} of \m{\prt_i}, 
is effectively a learned coarse grained representation of \m{\prt_i}. 
What makes physical problems different from, e.g., learning graphs, however is their spatial character. 
In particular: 
\begin{compactenum}[~1.]
\item Each subsystem \m{\Pcal_i} is now also 
associated with a vector \m{\V r_i\tin\RR^3} specifying its spatial position. 
\item The interaction between two subsystems \m{\prt_i} and \m{\prt_j} depends not only on 
their relative positions, but also on their relative orientation.  
Therefore, \m{\psi_i} and \m{\psi_j} must also have spatial character, 
somewhat similarly to the terms of the familiar monopole, dipole, quadrupole, etc.\:expansion.  
\end{compactenum}
If we rotate the entire the atomic environment around the central atom by some rotation \m{R\tin\SO(3)}\footnote{
\m{SO(3)} denotes the group of rotations in \m{\RR^3}, i.e., the group of three dimensional orthogonal, 
unit determinant matrices.}, the position vectors transform as \m{\V r_i\mapsto R\ts \V r_i}. 
Mathematically, the second point above says that the \m{\psi_i} activations (states) must also 
transform under rotations in a predictable way, 
which is expressed by saying that they must be rotationally \emph{covariant}.

\subsection{Group representations and \m{N}--body networks} 
Just as covariance to permutations is the critical constraint on the graph CCNs, 
covariance to rotations is the guiding principle behind CCNs for 
learning atomic force fields. To describe this concept in its general form, 
we start out by assuming only that any given activation \m{\psi} is representable 
as a \m{d} dimensional (complex valued) vector, and that the transformation 
that \m{\psi} undergoes under a rotation \m{R} is linear, i.e., \m{\psi\mapsto \rho(R)\ts\psi} 
for some matrix \m{\rho(R)}. 

The linearity assumption is sufficient to guarantee that for any \m{R,R'\tin\SO(3)}, 
\m{\rho(R)\ts\rho(R')\<=\rho(RR')}. 
Complex matrix valued functions satisfying this criterion are called 
\emph{representations} of the group \m{\SO(3)}. 
Standard theorems in representation theory 
tell us that any compact group \m{G} (such as \m{SO(3)}) has a sequence 
of so-called inequivalent irreducible representations \m{\seqzi{\rho}} (irreps, for short), and that 
any other representation \m{\mu} of \m{G} 
can be reduced into a direct sum of irreps in the sense that there is some 
invertible matrix \m{C} and sequence of integers \m{\seqzi{\tau}} such that 
\begin{equation}\label{eq: decomp1}
\mu(R)=C^{-1} \sqbBig{\bigoplus_\ell \bigoplus_{m=1}^{\tau_\ell} \rho_\ell(R)}\,C.
\end{equation}
Here \m{\tau_\ell} is called the \emph{multiplicity} of \m{\rho_\ell} in \m{\mu},  
and \m{\V \tau=(\seqzi{\tau})} is called the \emph{type} of \m{\mu}. 
Another nice feature of the representation theory of compact groups is that the irreps 
can always be chosen to be unitary, i.e., \m{\rho(R^{-1})=\rho(R)^{-1}=\rho(R)^\dag}, 
where \m{M^\dag} denotes the Hermitian conjugate (conjugate transpose) of the matrix \m{M}.  
In the following we will always assume that irreps satisfy this condition. 
If \m{\mu} is also unitary, then the transformation matrix \m{C} will be unitary too, 
so we can replace \m{C^{-1}} with \m{C^\dag}.  
For more background in representation theory, the reader is referred to \citep{Serre}. 

In the specific case of the rotation group \m{\SO(3)}, the irreps are sometimes called Wigner D--matrices. 
The \m{\ell\<=0} irrep consists of the one dimensional constant matrices \m{\rho_0(R)\<=(1)}, 
the \m{\ell\<=1} irrep (up to conjugation) is equivalent to the rotation matrices themselves, 
while for general \m{\ell}, assuming that \m{(\theta,\phi,\psi)} are the 
Euler angles of \m{R}, \m{[\rho_\ell(R)]_{m,m'}=e^{i\psi m'}\,Y^\ell_m(\theta,\phi)}, where 
\m{\cbrN{Y^\ell_m}} are the well known spherical harmonic functions. 
In general, the dimensionality of \m{\rho_\ell} is \m{2\ell\<+1}, i.e., 
\m{\rho_\ell(R)\tin \CC^{(2\ell+1)\times (2\ell+1)}}. 
 
\begin{definition}
We say that \m{\psi\tin\CC^d} is an  \df{SO(3)--covariant vector of type} \m{\V \tau=(\tau_0,\tau_1,\tau_2,\ldots)} 
if under the action of rotations it transforms as 
\begin{equation}\label{eq: covariant1}
\psi\mapsto \sqbBig{\bigoplus_\ell \bigoplus_{m=1}^{\tau_\ell} \rho_\ell(R)}\:\psi.
\end{equation}
Setting 
\begin{equation}\label{eq: covariant2}
\psi=\bigoplus_\ell \bigoplus_{m=1}^{\tau_\ell} \psi^\ell_m, 
\end{equation}
we call \m{\psi^\ell_m\nts\tin\CC^{2\ell+1}} the \m{(l,m)}\df{--fragment} of \m{\psi}, and 
\begin{equation*}
\psi^\ell= \bigoplus_{m=1}^{\tau_\ell} \psi^\ell_m
\end{equation*}  
the \m{\ell}'th \df{part} of \m{\psi}. 
A covariant vector of type \m{\V \tau=(0,0,\ldots,0,1)}, where the single \m{1} corresponds to \m{\tau_k}, 
we call an \df{irreducible vector of order k} or an \df{irreducible} \m{\mathbf{\rho_k}}\df{--vector}. 
Note that a first order irreducible vector is just a scalar. 
\end{definition}

\noindent 
The motivation behind the above definition is that each fragment \m{\psi^\ell_m} 
transforms in the very simple way \m{\psi^\ell_m\mapsto \rho_\ell(R)\,\psi^\ell_m}. 
Note that the words ``fragment'' and ``part'' are not standard in the literature, 
but we find them useful for describing covariant neural architectures. 
Also note that unlike \rf{eq: decomp1}, there is no matrix \m{C} in 
equations \rf{eq: covariant1} and \rf{eq: covariant2}. 
This is because if a given vector \m{\psi} transforms according to a general representation \m{\mu} 
whose decomposition does include a nontrivial \m{C}, this matrix can be easily be factored out 
by redefining \m{\psi} as \m{C\psi}. 
Here \m{\psi^\ell} is sometimes also called the projection of \m{\psi} to the \m{\ell}'th \emph{isotypic 
subspace} of the representation space that \m{\psi} lives in and \m{\psi=\psi^0\oplus\psi^1\oplus\ldots} 
is called the \emph{isotypic decomposition} of \m{\psi}. 
With these representation theoretic tools in hand, we define the concept of 
\m{\SO(3)}--covariant \m{N}--body neural networks as follows. 

\input{atomic1}
\begin{definition}\label{def: cnm}
Let \m{\mathcal S} be a physical system made up of \m{n} particles \m{\sseq{\xi}{n}}. 
An  \m{\SO(3)}--\df{covariant N--body neural network} \m{\Ncal} for \m{\Scal} is  
a composition scheme \m{\Dcal} in which 
\begin{compactenum}[~~1.]
\item Each node \m{\neuron_j}, which we will sometimes also call a \df{gate}, is associated with 
\begin{compactenum}
\item a physical sybsystem \m{\Pcal_j} of \m{\Scal}; 
\item a vector \m{\V r_j\tin\RR^3} describing the spatial poition of \m{\prt_j}; 
\item a vector \m{\psi_j} that that describes the internal state of \m{\prt_j} 
and is type \m{\V \tau_j} covariant to rotations. 
\end{compactenum}
\item If \m{\neuron_j} is a leaf node, then \m{\psi_j} is determined by the corresponding 
particle \m{\xi_j}. 
\item If \m{\neuron_j} is a non-leaf node and its children are \m{\neuron_{\ch_1},\ldots,\neuron_{\ch_k}}, 
then \m{\psi_j} is computed as 
\begin{equation}\label{eq: aggregation}
\psi_j=\Phi_j(\h r_{\!\ch_1},\ldots,\h r_{\ch_k},
\h {\V r}_{\!\ch_1},\ldots,\h {\V r}_{\!\ch_k},\psi_{\ch_1}\ldots,\psi_{\ch_k}), 
\end{equation}
where \m{\h {\V r}_{\!\ch_{i}}=\V r_{\!\ch_i}\<-\V r_{\!j}} and \m{\h r_i=\abs{\h{\V r}_i}}. 
We call \m{\Phi_j} the local \df{aggregation rule}. 
\item \m{\Dcal} has a unique root \m{\neuron_r}, and the output of the network, 
i.e., the learned state of the entire system is \m{\psi_r}. 
In the case of learning scalar valued functions, such as the atomic potential, \m{\psi_r} is just a scalar. 

\end{compactenum}
\end{definition}

\noindent 
Note that what is described in Definition \ref{def: cnm} is a general architecture for learning 
the state of \m{N}--body physical systems with much wider applicability than just learning atomic potentials. 
The main technical challenge of the present paper is to define the \m{\Phi_j} aggregation rules 
in such a way as to guarantee that each \m{\psi_j} is \m{\SO(3)}--covariant. 
This is what is addressed in the following section. 

%% file: atomic1.tex
\begin{figure}[t!]
\includepic{.4}{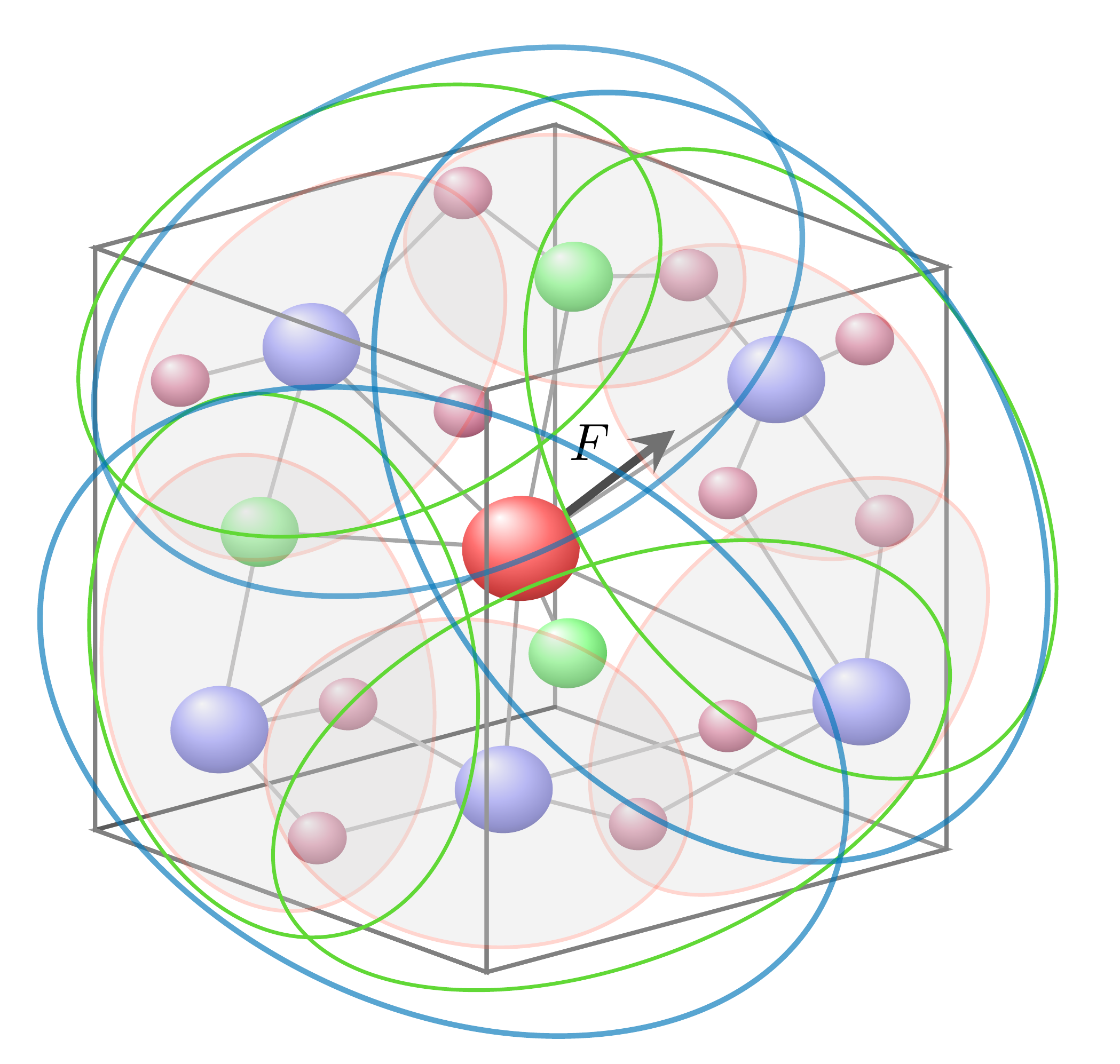}
\hspace{40pt}
\includepic{.3}{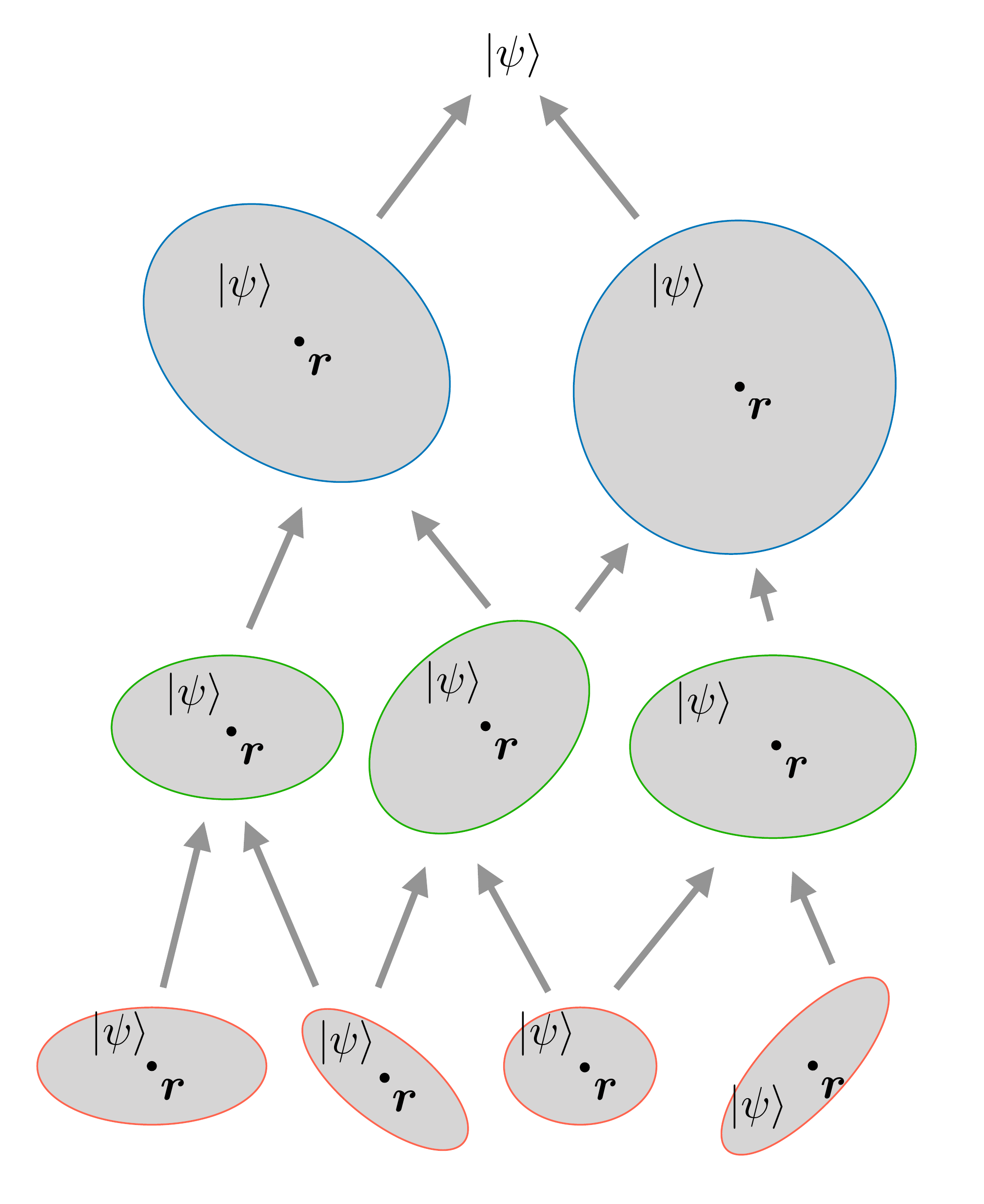}
\caption{\label{fig: atomic1}
In a comp-net for learning atomic force fields, the output of each ``part'' \m{\Pcal_i} is 
\m{(\V r_i, \psi_i)}, where \m{\V r_i} is the position vector of the corresponding physical subsystem, 
and \m{\psi_i} is a vector describing its internal state. 
}
\end{figure}

%% file: activations.tex
\section{Covariant aggregation rules}


To define the aggregation function \m{\Phi} to be used in \m{\SO(3)}--covariant comp-nets, 
all that we assume is that it is a polynomial in the relative positions  
\m{\h{\V r}_{\ch_1},\ldots,\h{\V r}_{\ch_k}}, the constituent state vectors 
\m{\psi_{\ch_1},\ldots,\psi_{\ch_k}} and the inverse distances 
\m{1/\h{r}_{\ch_1},\ldots 1/\h{r}_{\ch_k}}. 
Specifically, we say that \m{\Phi} is a \m{(P,Q,S)}--order aggregation function if each component of 
\m{\psi=\Phi(\h r_{\ch_1},\ldots,\h r_{\ch_k},\h{\V r}_{\ch_1},\ldots,\h{\V r}_{\ch_k},\psi_{\ch_1}\ldots,\psi_{\ch_k})} is a 
polynomial of order at most \m{p} in each component of \m{\V r_{\ch_i}}, a polynomial 
of at most \m{q} in each component of \m{\psi_{\ch_i}}, and a polynomial of 
order at most \m{s} in each \m{1/(\h r_{\ch_i})}. 
Any such \m{\Phi} can be expressed as 
\begin{equation}\label{eq: aggreg1}
\Phi(\ldots)=\mathcal{L}\brBig{\bigoplus_{\mathbf{p},\,\mathbf{q},\, \mathbf{s}} 
\V r_{\ch_1}^{\otimes p_1}\otimes \ldots\otimes \V r_{\ch_k}^{\otimes p_k}\otimes 
\psi_{\ch_1}^{\otimes q_1}\otimes\ldots\otimes \psi_{\ch_k}^{\otimes q_k}
\cdot \h r_{\ch_1}^{-s_1}\cdot\ldots\cdot \h r_{\ch_k}^{-s_k}}, 
\end{equation}
where \m{\mathbf{p}, \mathbf{q}} and \m{\mathbf{s}} are multi--indices of positive integers 
with \m{p_i\<\leq P},\: \m{q_i\<\leq Q} and \m{s_i\<\leq S}, and \m{\mathcal{L}} is a linear function. 
The tensor products appearing in \rf{eq: aggreg1} are formidably large object that in most cases 
would be impractical to compute explicitly. 
Rather, this equation is just meant to emphasize that 
any learnable parameters of the network must be implicit in the linear operator \m{\mathcal{L}}. 

The more stringent requirements on \m{\mathcal{L}} arise from the covariance criterion. 
The key to understanding these is the observation that for any 
sequence \m{\sseq{\rho}{p}} of (not necessarily irreducible) representations of a 
compact group \m{G}, their tensor product 
\[\rho(R)=\rho_1(R)\otimes \rho_2(R)\otimes\ldots\otimes \rho_p(R)\] 
is also a representation of \m{G}. 
Consequently, \m{\rho} has a decomposition into irreps, similar to \rf{eq: decomp1}. 
As an immediate corollary, any product of \m{\SO(3)} covariant vectors can be similarly decomposed.  
In particular, by applying the appropriate unitary matrix \m{C},  
the sum of tensor products appearing in \rf{eq: aggreg1} can be decomposed into a sum of 
irreducible fragments in the form 
\begin{equation*}
\bigoplus_{\ell=0}^L \bigoplus_{m=1}^{\tau'_\ell} \phi^\ell_m=C\brbigg{
\bigoplus_{\mathbf{p},\,\mathbf{q},\, \mathbf{s}} 
\V r_{\ch_1}^{\otimes p_1}\otimes \ldots\otimes \V r_{\ch_k}^{\otimes p_k}\otimes 
\psi_{\ch_1}^{\otimes q_1}\otimes\ldots\otimes \psi_{\ch_k}^{\otimes q_k}
\cdot \h r_{\ch_1}^{-s_1}\cdot\ldots\cdot \h r_{\ch_k}^{-s_k}}. 
\end{equation*}
To be explicit, we define 
\begin{equation}\label{eq: projections}
\phi^\ell_m=T^\ell_m\brbigg{
\bigoplus_{\mathbf{p},\,\mathbf{q},\, \mathbf{s}} 
\V r_{\ch_1}^{\otimes p_1}\otimes \ldots\otimes \V r_{\ch_k}^{\otimes p_k}\otimes 
\psi_{\ch_1}^{\otimes q_1}\otimes\ldots\otimes \psi_{\ch_k}^{\otimes q_k}
\cdot \h r_{\ch_1}^{-s_1}\cdot\ldots\cdot \h r_{\ch_k}^{-s_k}},
\end{equation}
where \m{T^0_1,\ldots,T^0_{\tau_0},T^1_1,\ldots,T^1_{\tau_2},\ldots,T^L_{\tau_L}} is an appropriate 
sequence of projection operators. 
The following proposition is a key result of our paper. 

\begin{proposition}\label{prop: aggreg}
The output of the aggregation function \rf{eq: aggreg1} is a \m{\V \tau}--covariant vector 
if and only if \m{\mathcal{L}} is of the form 
\begin{equation}\label{eq: weight1}
\mathcal{L}(\ldots)= 
\bigoplus_{\ell=0}^L \bigoplus_{m=1}^{\tau_\ell} \sum_{m'=1}^{\tau'_\ell}
w^\ell_{m',m}\, \phi^\ell_{m'}. 
\end{equation}
Equivalently, collecting all \m{\phi^\ell_{m'}} fragments with the same \m{\ell} into a matrix 
\sm{\tilde F^\ell\tin \CC^{(2\ell\<+1)\times \tau'_\ell}}, all \m{(w^\ell_{m',m})_{m',m}} 
weights into a matrix \sm{W^\ell\tin \CC^{\tau'_\ell\<\times \tau_\ell}}, 
and reinterpreting the output of \m{\mathcal{L}} as a collection of matrices rather than a 
single long vector,  
\begin{equation}\label{eq: weight2}
\mathcal{L}(\ldots)= 
\brbig{\tilde F^0 W^0, \tilde F^1 W^1,\ldots,\tilde F^L W^L}. 
\end{equation}
\end{proposition}

\noindent 
Proposition \ref{prop: aggreg} tell us that \m{\Lcal} is only allowed to mix \m{\phi^\ell_m} fragments 
with the same \m{\ell}, and that fragments can only be mixed in their entirety, rather than picking 
out their individual components. These are crucial consequences of equivariance. 
However, there are no further restrictions on the \m{\brN{W^\ell}_\ell} mixing matrices. 

In an \m{N}--body neural network the \m{W^\ell} matrices are shared across (some subsets of) 
nodes, and it is these mixing (weight) matrices that the network learns from training data. 
The \sm{\tilde F^\ell} matrices can be regarded as generalized matrix valued activations. 
Since each \m{W^\ell} interacts with the \m{F^\ell} matrices linearly, the network can be trained the usual 
way by backpropagating gradients of whatever loss function is applied to the output node 
\m{\neuron_r}, whose activation is usually scalar valued. 
 
It is important to note that \m{N}--body neural networks have no additional nonlinearity outside 
of \m{\Phi}, since that would break covariance. 
In contrast, in most existing neural network architectures, as explained in Section \ref{sec: nn}, 
each neuron first takes a linear combination of its inputs weighted by learned weights and then 
applies a fixed pointwise nonlinearity, \m{\sigma}. 
In our architecture the nonlinearity is hidden in the way that the 
\m{\phi^\ell_m} fragments are computed, since a tensor product is a nonlinear function of its factors. 
On the other hand, mixing the resulting fragments with the \m{W^\ell} weight matrices is a linear operation. 
Thus, in our case, the nonlinear part of the operation \emph{precedes} the linear part. 

The generic polynomial aggregation function \rf{eq: aggreg1} is too general to be used in a practical 
\m{N}--body network, and would be far too costly computationally. 
Instead, we propose using a few specific types of low order gates, such as those described below. 

\subsubsection{Zeroth order interaction gates}\label{sec: zeroth order}

Zeroth order interaction gates aggregate the states of their children and combine them with their relative 
position vectors, but do not capture interactions between the children. 
A simple example of such a gate would be one where 
\begin{equation}\label{eq: 0th order gate}
\Phi(\ldots)=\mathcal{L}\brBig{\:\Sum{i}{k}\ts (\psi_{\ch_i}\<\otimes \h{\V r}_{\ch_i}),\;  
\Sum{i}{k} \h r_{\ch_i}^{-1}\ts(\psi_{\ch_i}\<\otimes \h{\V r}_{\ch_i}),\; 
\Sum{i}{k} \h r_{\ch_i}^{-2}\ts(\psi_{\ch_i}\<\otimes \h{\V r}_{\ch_i})\:}.
\end{equation}
Note that the summations in these formulae ensure that the output is invariant with respect to 
permuting the children and also reduce the generality of \rf{eq: aggreg1} because the direct 
sum is replaced by an explicit summation (this can also be interpreted as tying some of the 
mixing weights together in a particular way). 
Let \m{L} be the largest \m{\ell} for which \m{\tau_\ell\neq 0} in the inputs.  
In the \m{L\<=0} case each \m{\psi_{\ch_i}} state is a scalar quantity, such as electric charge. 
In the \m{L\<=1} case it is a vector, such as the dipole moment. 
In the \m{L\<=2} case it can encode the quadropole moment, and so on. 
A gate of the above form can learn how to combine such moments into a single (higher order) moment 
corresponding to the parent system. 

It is instructive to see how many parameters a gate of this type has. 
Let us assume the simple case that each \m{\psi_{\ch_i}} is of type \m{\V\tau\<=(1,1,\ldots,1)} 
(up to \m{\ell\<=L}).  
The type of \m{\h{\V r}_{\ch_i}} is \m{(0,1)}. 
According to the Clebsch--Gordan rules (see Section \ref{sec: CG}), the product of two such vectors is 
a vector of type \m{(1,3,2,\ldots,2,1)} (of length \m{L+1}).  
Further assume that desired output type is again \m{\V\tau\<=(1,1,\ldots,1)} of length \m{L}. 
This means that the \m{\ell=L\<+1} fragment does not even have to be computed, and the 
size of the weight matrices appearing in \rf{eq: weight2} are 
\[W_0\in \CC^{1\times 3}\qquad W_1\tin \CC^{1\times 9} \qquad W_2\tin \CC^{1\times 6}\qquad \ldots\qquad  
W_L\tin \CC^{1\times 6}.
\]
The size of these matrices changes dramatically as we allow more ``channels''. 
For example, if each of the input states are of type \m{\V\tau\<=(c,c,\ldots,c)}, the type of 
\m{\psi_{\ch_i}\nts\otimes \h{\V r}_{\ch_i}} becomes \m{(c,3c,2c,\ldots,2c,1c)}. 
Assuming again an output of type \m{\V\tau\<=(c,c,\ldots,c)}, the weight matrices become 
\[W_0\in \CC^{c\times 3c}\qquad W_1\tin \CC^{c\times 9c} \qquad W_2\tin \CC^{c\times 6c}\qquad \ldots\qquad  
W_L\tin \CC^{c\times 6c}.
\]
In many networks, however, the number of channels increases as we go higher in the network. 
Allowing the output type to be as rich as possible, without inducing linear redundancies, 
the output type becomes \m{(3c,9c,6c,\ldots,6c,3c)}, and 
\[W_0\in \CC^{3c\times 3c}\qquad W_1\tin \CC^{9c\times 9c} \qquad W_2\tin \CC^{6c\times 6c}\qquad \ldots\qquad  
W_L\tin \CC^{6c\times 6c}.
\]

\subsubsection{First order interaction gates}

In first order interaction gates each of the children interact with each other, and the parent 
aggregates these pairwise interactions. A simple example would be computing the total energy of a 
collection of charged bodies, which might be done with a gate of the form
\begin{multline}\label{eq: first order gate}
\Phi(\ldots)=\mathcal{L}\brBig{\:
\Sum{i,j}{k}\ts (\psi_{\ch_i}\<\otimes \psi_{\ch_j}\<\otimes\h{\V r}_{\ch_i}\<\otimes \h{\V r}_{\ch_i}),\;  
\Sum{i,j}{k}\h r_{\ch_i}^{-1}\h r_{\ch_j}^{-1}\ts (\psi_{\ch_i}\<\otimes \psi_{\ch_j}\<\otimes\h{\V r}_{\ch_i}\<\otimes \h{\V r}_{\ch_j}),\;\\  
\Sum{i,j}{k}\h r_{\ch_i}^{-2}\h r_{\ch_j}^{-2}\ts (\psi_{\ch_i}\<\otimes \psi_{\ch_j}\<\otimes\h{\V r}_{\ch_i}\<\otimes \h{\V r}_{\ch_j}),\;  
\Sum{i,j}{k}\h r_{\ch_i}^{-3}\h r_{\ch_j}^{-3}\ts (\psi_{\ch_i}\<\otimes \psi_{\ch_j}\<\otimes\h{\V r}_{\ch_i}\<\otimes \h{\V r}_{\ch_j})\;  
}.
\end{multline}
Generalizing \rf{eq: aggreg1} slightly, if we know that the interaction only depends on the relative positions 
of the child systems, we can also use 
\begin{multline}\label{eq: first order gate}
\Phi(\ldots)=\mathcal{L}\brBig{\:
\Sum{i,j}{k}\ts (\psi_{\ch_i}\<\otimes \psi_{\ch_j}\<\otimes\h{\V r}_{\ch_i,\ch_j}),\;  
\Sum{i,j}{k}\h r_{\ch_i,\ch_j}^{-1}\ts (\psi_{\ch_i}\<\otimes \psi_{\ch_j}\<\otimes\h{\V r}_{\ch_i,\ch_j}),\;\\  
\Sum{i,j}{k}\h r_{\ch_i,\ch_j}^{-2}\ts (\psi_{\ch_i}\<\otimes \psi_{\ch_j}\<\otimes\h{\V r}_{\ch_i,\ch_j}),\;  
\Sum{i,j}{k}\h r_{\ch_i,\ch_j}^{-3}\ts (\psi_{\ch_i}\<\otimes \psi_{\ch_j}\<\otimes\h{\V r}_{\ch_i,\ch_j})\;  
},
\end{multline}
where \m{\h{\V r}_{\ch_i,\ch_j}=\h{\V r}_{\ch_i}\!-\h{\V r}_{\ch_j}} and 
\m{\h r_{\ch_i,\ch_j}=\absN{\h{\V r}_{\ch_i,\ch_j}}}. 

It is important to note that in the above electrostatics was used only as an example. 
There is no need to learn electrostatic interactions because they are perfectly described by classical 
physics. Rather, we envisage using the zeroth and first order interaction gates as constituents of a larger  
network for learning more complicated interactions with no simple closed form that nonetheless 
broadly follow similar scaling laws as classical interactions. 

\subsection{Clebsch--Gordan transforms}\label{sec: CG}

It remains to explain how the \m{T^\ell_m} projection maps appearing in \rf{eq: projections} are 
computed. This is critical because the nonlinearities in our network are the tensor products, 
and our architecture hinges on being able to reduce vectors into a direct sum of irreducibles again 
straight after the tensor product operation. 

Fortunately, representation theory provides a clear prescription for how this operation is to be performed. 
For any compact group \m{G}, given two irreducible representations \m{\rho_{\ell_1}} and \m{\rho_{\ell_2}}, 
the decomposition of \m{\rho_{\ell_1}\otimes \rho_{\ell_2}} into a direct sum of irreducibles 
\begin{equation}\label{eq: CG1}
\rho_{\ell_1}\nts(R)\otimes \rho_{\ell_2}\nts(R)=
C_{\ell_1\nts,\ell_2}^\dag \sqbBig{\:\bigoplus_{\ell}\bigoplus_{m=1}^{\kappa_{{{\tau}_1\nts,\mathbf \tau_2}}\nts(\ell)~} 
\!\!\rho_\ell(R)\:}\, C_{\ell_1\nts,\ell_2}
\end{equation}
is called the Clebsch--Gordan transform. 
In the specific case of \m{\SO(3)}, the \m{\kappa} multiplicities take on the very simple form 
(which we already used in Section \ref{sec: zeroth order}) 
\[\kappa_{\ell_1\nts,\ell_2}(\ell)=
\begin{cases}
~1&\text{if}~~\abs{\ell_1\<-\ell_2}\leq \ell\leq \ell_1\<+\ell_2\\
~0&\Totherwise,
\end{cases}\]
and the elements of the \m{C_{\ell_1\nts,\ell_2}} matrices can also be computed relatively 
easily via closed form formulae. 

We immediately see that \rf{eq: CG1} tells us how to reduce the product of covariant vectors into irreducible 
fragments. Assuming for example that \m{\psi_1} is an irreducible \m{\rho_{\ell_1}} vector and \m{\psi_2} is an 
irreducible \m{\rho_{\ell_2}} vector, \m{\psi_1\otimes \psi_2} decomposes into irreducible fragments in the form 
\[
\psi_1\otimes\psi_2= \bigoplus_{\ell=\abs{\ell_1\<-\ell_2}}^{\ell_1\<+\ell_2} \wbar \psi{}^\ell 
\qqquad\text{where}\qqquad \wbar{\psi}{}^\ell=C_{\ell_1\nts,\ell_2\nts,\ell}\,(\psi_1\otimes\psi_2), 
\]
and \m{C_{\ell_1\nts,\ell_2\nts,\ell}} is the part of  \m{C_{\ell_1\nts,\ell_2}} matrix 
corresponding to the \m{\ell}'th ``block''. 
Thus, in this case the operator \m{T^\ell_1} just corresponds to mutiplying the tensor product by  
\m{C_{\ell_1\nts,\ell_2\nts,\ell}}.
By linearity, the above relationship also extends to non-irreducible vectors. 
If \m{\psi_1} is of type \m{\V \tau_1} and \m{\psi_2} is of type \m{\V \tau_2}, then 
\[
\psi_1\otimes\psi_2= \bigoplus_\ell \bigoplus_{m=1}^{\kappa_{\tau_1\nts,\tau_2}(\ell)} \wbar{\psi}{}^\ell_m
\]
where 
\[
\kappa_{\V \tau_1\nts,\V \tau_2}(\ell)=\sum_{\ell_1}\sum_{\ell_2}\; [\tau_1]_{\ell_1} \cdot[\tau_2]_{\ell_2}\:
\cdot \II\sqb{\abs{\ell_1\<-\ell_2}\leq \ell\leq \ell_1\<+\ell_2}, 
\]
and \m{\II[\cdot]} is the indicator function. Once again, the actual \m{\wbar{\psi}{}^\ell_m} fragments 
are computed by applying the appropriate \m{C_{\ell_1\nts,\ell_2\nts,\ell}} matrix to the 
appropriate combination of irreducible fragments of \m{\psi_1} and \m{\psi_2}. 
It is also clear that the by applying the Clebsch--Gordan decomposition recurisively, we can 
decompose a tensor product of any order, e.g., 
\[
\psi_1\otimes\psi_2\otimes \psi_3\otimes \ldots\otimes \psi_k=
((\psi_1\otimes\psi_2)\otimes \psi_3)\otimes \ldots\otimes \psi_k.
\]
In an actual computation of such higher order products, however, a considerable amount of thought might 
have to go into optimizing the order of operations and reusing potential intermediate results to minimize 
computational cost.

\ignore{
can decompose higher order products 

As a direct consequence, given 

By linearity 

\begin{proposition}
Let \m{\V v_1} and \m{\V v_2} be vectors of type \m{\V \tau_1} and \m{\V \tau_2} with respect to the 
action of \m{SO(3)}. Then \m{\V v_1\<\otimes \V v_2} is a vector of type \m{\V \tau}, where 
\[\tau_\ell=\sum_{\ell_1} \sum_{\ell_2} \kappa_{\tau_1,\tau_2}(\ell),\]
and decomposes into irreducible covariant vectors in the form 
\begin{equation}\label{eq: CG2}
\V v^{\ell_1,\ell_2,\ell}_m=C_{\ell_1,\ell_2,\ell,m}\brbig{[\V v_1]_{\ell_1} \otimes [\V v_2]_{\ell_2}}. 
\end{equation}
\end{proposition}
\bigskip

\noindent
The tensor product appearing in \rf{eq: aggreg2} could, in principle be decomposed by repeated application 
of \rf{eq: CG2} to 
\m{\V r_{\ch_1} \<\otimes \V r_{\ch_1}}, 
\m{(\V r_{\ch_1} \<\otimes \V r_{\ch_1})\<\otimes \V r_{\ch_1}}, and so on. 
In the general case, however, this would quickly get very expensive.  

Therefore, instead of the general tensor product, we use the reduced form 
\[\sum_{i=1}^{k} \sum_{j=1}^k \V r_{\ch_i}\otimes \V r_{\ch_j}\otimes \psi_{\ch_i}\otimes \psi_{\ch_j}.\]
}

\ignore{
We say that \m{\Phi} is \emph{irreducible} if 
there is no linear dependence between the components of \m{\psi} as 
the \m{\cbrN{\V r_{\ch_i}}} and \m{\cbrN{\psi_{\ch_i}}} argumets vary. 

\begin{proposition}\label{prop: aggreg1}
Any irreducible \m{(p,q)}--order aggregation function can be written in the form 
\begin{equation}\label{eq: aggreg1}
\Phi(\ldots)=T\brbig{\V r_{\ch_1}^{\otimes p}\otimes \ldots\otimes \V r_{\ch_k}^{\otimes p}\otimes 
\psi_{\ch_1}^{\otimes q}\otimes\ldots\otimes \psi_{\ch_k}^{\otimes q}}, 
\end{equation}
where \m{\V v^{\otimes p}} denotes the \m{p}--fold tensor product \m{\V v\otimes \V v\otimes\ldots}, 
and \m{T} is a linear transformation \m{\CC^{n}\to\CC^m} with \m{m\<\leq n}. 
\end{proposition}
\bigskip 
}
\ignore{
The following proposition is an immediate corollary. 

\begin{proposition}\label{prop: decomp}
Let \m{\sseq{\V v}{k}} be a sequence of vectors of types \m{\sseq{\V \tau}{k}} with respect to the 
action of \m{\SO(3)}. Then there is a type \m{\V \tau} and a sequence of linear transformations 
\m{\cbrN{T^\ell_m}} such that each 
\[\phi^\ell_m:=T^\ell_m(\V v_1\otimes\V v_2\otimes\ldots\otimes \V v_k)\qquad \ell=0,1,2,\ldots,\qquad 
m=1,2,\ldots \V \tau_\ell\]
is an irreducible \m{\rho_\ell} vector, and \m{\bigoplus_\ell \bigoplus_m T^\ell_m} is a unitary
transformation. 
\end{proposition}
\noindent
}

\ignore{
The \m{W_\ell=(w^{\ell}_{m,m'})_{m,m'}} matrices contain the weights the are learnt by the neural network. 
Since \rf{eq: aggreg2} is linear in the components of the tensor product 
\m{\V r_{\ch_1}^{\otimes p}\otimes\ldots  
\otimes \psi_{\ch_k}^{\otimes q}} as well as in the \m{W_\ell} weight matrices, 
the components of the weight matrices can, in principle, be learnt the same way as it is done 
in more customary feed forward neural nets. 
It remains to derive the algorithm used to compute the \m{T^\ell_{m'}} transforms. 
}

%% file: conclusions.tex
\section{Conclusions}

There is considerable excitement in both the Machine Learning and the Physics/Chemistry 
communities about the potential of using neural networks to learn to the behavior and properties of 
complex physical systems. However, phyiscal systems have nontrivial invariance properties (in particular, 
invariance to translations, rotations and the exchange of identical elementary parts) that must be 
strictly respected. 

In this paper we proposed a new type of generalized convolutional neural network architecture, 
\m{N}\emph{--body networks}, 
which provides a flexible framework for modeling interacting systems of various types, 
while taking into account these invariances (symmetries). 
The specific motivation for developing \m{N}\emph{--body networks} is to learn atomic potentials 
(force fields) for molecular dynamics simulations. However, we envisage that they will be used 
more broadly, for modeling a variety of systems. 
The closest to our work in certai ways are Moment Tensor Potientials \citep{Shapeev2015}, 
although that framework does not have learnable parameters. 

\m{N}--body networks are distinguished from earlier neural network models for physical systems in that 
\begin{compactenum}[~~1.]
\item The model is based on a hierarchical (but not necessarily strictly tree-like) decomposition of the system into 
subsystems at different levels, which is directly reflected in the structure of the neural network. 
\item Each subsystem is identified with a ``neuron'' (or ``gate'') \m{\neuron_i} in the network, 
and the output (activation) \m{\psi_i} of the neuron becomes a representation of the subsystem's internal state. 
\item The \m{\psi_i} states are tensorial objects with spatial character, in particular they are \emph{covariant} 
with rotations in the sense that they transform 
under rotations according to specific irreducible representations of the rotation group. 
The gates are specially constructed to ensure that this covariance property is preserved throught the network.  
\item Unlike most other neural network architectures, the nonlinearities in \m{N}--body networks are not 
pointwise operations, but are applied in ``Fourier space'', i.e., directly to the irreducible parts 
of the state vector objects. This is only possible because (a) the nonlinearities arise as a consequence 
of taking tensor products of covariant objects (b) the tensor products are decomposed into irreducible 
parts by the Clebsch--Gordan transform.  
\end{compactenum}
We believe that the last of these ideas is particularly promising, because it suggests the possibility of 
constructing neural that operate entirely in Fourier space, and use tensor products combined with 
Clebsch--Gordan transforms to induce nonlinearities. This might have significance 
for a range of other applications, as well. 
Experiments are ongoing to validate our framework on real physical systems. 

%% file: acknowledgements.tex
\subsection*{Acknowledgements}

The author would like to thank Shubhendu Trivedi, Brandon Anderson, Hy Truong Son, Horace Pan, G\'abor 
Cs\'anyi and Michele Ceriotti for their input to this work. 
Financial support for this work was provided in part by DARPA award number D16AP00112.  